\documentclass[11pt]{article}

\usepackage{acl}

\usepackage{times}
\usepackage{latexsym}

\usepackage[T1]{fontenc}

\usepackage[utf8]{inputenc}
\usepackage{dsfont}
\usepackage{pifont}
\usepackage{bbm}    
\usepackage{xcolor} 
\usepackage{microtype}

\usepackage{inconsolata}

\usepackage{tabularx}

\usepackage[most]{tcolorbox}
\usepackage{listings}
\tcbuselibrary{listings} 

\usepackage{booktabs}
\usepackage{multirow}
\usepackage[table]{xcolor} 
\usepackage{graphicx}
\usepackage{amsmath} 
\usepackage{amssymb} 
\usepackage{hyperref}

\newtcolorbox{paperbox}[1]{
    colback=white, 
    colframe=gray!60, 
    arc=0pt, 
    outer arc=0pt,
    title=#1, 
    colbacktitle=gray!15, 
    coltitle=black,
    fonttitle=\bfseries\small,
    boxrule=0.6pt,
    left=10pt, right=10pt, top=8pt, bottom=8pt,
    width=\textwidth,
    after skip=\abovecaptionskip
}

\definecolor{MyGreen}{HTML}{70AD47}
\title{AdaTIR: Adaptive Tool-Integrated Reasoning via Difficulty-Aware Policy Optimization}

\author{Zhaiyu Fang\thanks{\,Corresponding author.}, ~Ruipeng Sun \\
  Trip.com Group \\
  \texttt{\{zhaiyufang, sunrp\}@trip.com} \\}

\begin{document}
\maketitle
\begin{abstract}

  Tool-Integrated Reasoning (TIR) has significantly enhanced the capabilities of Large Language Models (LLMs), yet current agents tend to exhibit cognitive offloading, redundantly invoking external tools even for simple tasks. In this paper, we suggest that true agentic intelligence requires not just tool invocation, but the adaptive wisdom to discern when to use them. We propose AdaTIR, a framework that shifts the paradigm from static tool invocation to difficulty-aware reasoning internalization. By introducing a difficulty-aware efficiency reward, AdaTIR dynamically adjusts tool budgets based on task complexity—internalizing reasoning for simple tasks while selectively invoking tools for complex tasks. Furthermore, we identify a sign reversal problem where tool penalties outweigh correctness rewards, mistakenly penalizing correct rollouts with negative advantages. To resolve this, we propose Clipped Advantage Shaping (CAS), which ensures that correctness remains the primary objective while using efficiency as a secondary constraint. Empirical results demonstrate that AdaTIR reduces tool calls by up to 97.6\% on simple tasks and 28.2\% on complex challenges while maintaining or enhancing accuracy. Notably, AdaTIR successfully internalizes reasoning, outperforming baselines by 4.8\% on AIME 2024 even when tool access is strictly disabled.

\end{abstract}

\section{Introduction}

\begin{figure}[t]
  \centering
  \includegraphics[width=0.8\columnwidth]{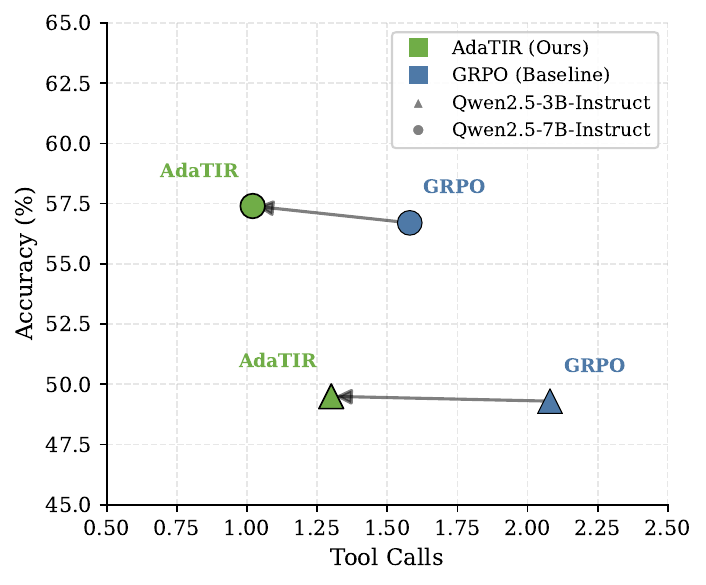}
  \caption{\textbf{Accuracy versus Tool Calls for Qwen2.5 under different training strategies.} GRPO and AdaTIR denote the baseline and our proposed method, respectively. AdaTIR consistently achieves a superior efficiency-accuracy trade-off across both 3B and 7B backbones by internalizing reasoning capabilities.}
  \label{fig:teaser_main}
\end{figure}

Large Language Models (LLMs)—such as OpenAI-o1~\citep{o1} and DeepSeek-R1~\citep{r1}—have achieved remarkable milestones in complex reasoning, manifesting superior capabilities in tasks ranging from mathematical problem-solving~\citep{aime} to repository-level code generation~\citep{swe} and open-domain question answering~\citep{nq}. To further transcend the limitations of parametric knowledge, \textbf{Tool-Integrated Reasoning (TIR)} has emerged as a powerful paradigm~\citep{tora, torl, retool, artist, simpletir}, augmenting Large Language Models (LLMs) with external executors to tackle problems intractable for pure-text generation~\citep{UTIR}. However, this augmentation comes with a hidden cognitive cost, current agents exhibit a pathological tendency toward ``cognitive offloading''~\citep{OTC, xu2024cognitive}. By frequently invoking tools for simple tasks, they not only incur unnecessary latency but also risk degrading their intrinsic reasoning consistency through environmental noise. The fundamental challenge lies in selectively internalizing tool-independent logic without compromising the robustness needed for genuinely complex tasks.

To address these efficiency bottlenecks, recent research has explored ranging from inference-time budget constraints~\citep{BATS} to training-time reward shaping. However, these approaches often fail to address the root cause of tool dependency. Inference-time constraints function as shallow prompt-level limits, often leading to context drift or reasoning dead-ends when the model lacks the internalized logic to proceed without tools~\citep{lost_in_the_mid, reflexion}. Alternatively, training-based methods like OTC impose indiscriminate penalties on tool invocation. This creates a fundamental dilemma: a penalty strong enough to suppress redundancy on simple tasks often inadvertently impairs the model's ability to solve complex tasks that genuinely require external assistance. Furthermore, approximating optimal budgets based on biased group-minima estimates often leads to sub-optimal convergence where tool redundancy persists.

To address these limitations, we propose \textbf{AdaTIR}, a framework that establishes a difficulty-aware policy to adaptively regulate tool invocation frequency based on task complexity. Our core insight is that tool invocation is inherently difficulty-dependent. Different from static penalties that indiscriminately suppress actions, we suggest that constraints should be dynamic: stricter budgets on simple tasks compel the model to internalize reasoning capabilities and mitigate ``cognitive offloading'', while relaxed constraints on complex tasks ensure the model retains the robustness to leverage external tools when necessary. Consequently, this paradigm enables the model to align computational expenditure with task difficulty, effectively pushing the cost-performance Pareto frontier.

The implementation of this strategy, however, reveals a critical algorithmic challenge: naive reward shaping inherently risks sign reversal, where efficiency penalties can inadvertently outweigh correctness rewards. Specifically, as training progresses, rollouts within a group exhibit consistent correctness. This phenomenon dilutes the accuracy-based advantage signal, allowing the auxiliary efficiency signal to disproportionately influence the optimization objective. Consequently, even rollouts that yield correct answers could be assigned a negative advantage if their tool invocation exceeds the group average, creating an objective misalignment that suppresses valid reasoning paths and induces training instability. To address this, we propose \textbf{Clipped Advantage Shaping (CAS)}, a mechanism that directly reformulates the advantage term rather than modifying the scalar reward. By integrating a clipped efficiency penalty as an auxiliary signal, CAS effectively encourages the prioritization of correctness signals over efficiency signals, causing the model to optimize efficiency without suffering from the mode collapse~\citep{entropy, entropy2} observed in conventional methods.

Our contributions are summarized as follows:
\begin{itemize}
  \item We propose AdaTIR, a framework that establishes a difficulty-aware policy to adaptively regulate tool invocations tailored to task complexity, effectively suppressing redundant "cognitive offloading" on simple tasks. 
  \item We introduce Clipped Advantage Shaping (CAS), a mechanism that mathematically reformulates the advantage term to maintain sign consistency, resolving the training instability and mode collapse often observed in naive reward shaping.
    
  \item We demonstrate that our approach significantly enhances efficiency across varying difficulty levels, reducing tool calls by up to 97.6\% (Figure~\ref{fig:teaser_main}) on simple tasks while maintaining robust performance on complex challenges. Notably, our method exhibits superior reasoning capabilities even when tool access is strictly disabled, indicating successful reasoning internalization.
  
\end{itemize}

\begin{figure*}[t] 
  \centering
  \includegraphics[width=0.98\textwidth]{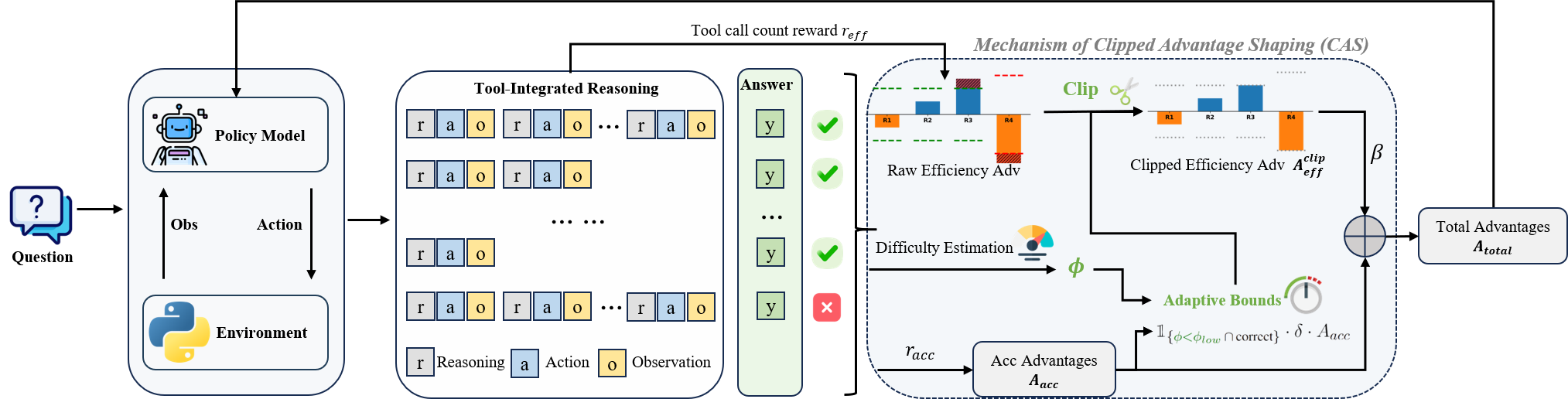} 
  \caption{\textbf{Overview of the proposed framework.} 
  Efficiency signals are conditionally injected only for correct rollouts on \textcolor{MyGreen}{\textbf{easy}} tasks (where difficulty estimation $\phi$ falls below threshold $\phi_{low}$). 
  These auxiliary signals are explicitly clipped by $\delta |A_{acc}|$ ($0 < \delta < 1$), to ensure efficiency optimization remains secondary to correctness, thereby maintaining sign consistency.}
  \label{fig:main_framework}
\end{figure*}

\section{Preliminaries}
\textbf{Tool-Integrated Reasoning (TIR).} 
In the TIR paradigm, the large language model acts as a reasoning engine that interleaves text generation with external tool interactions~\citep{mathcoder, deepseek_math}. Given a task $q$, the reasoning process of the policy $\pi_\theta$ forms a trajectory $\tau$ consisting of sequential reasoning steps and environment feedbacks:
\begin{equation}
    \tau = (w_1, c_1, o_1, w_2, c_2, o_2, \dots, w_K, c_K, o_K, \hat{y})
\end{equation}
where $w_k$ represents the $k$-th reasoning thought, $c_k$ is the tool invocation command, and $o_k$ is the corresponding observation returned by code execution.

\textbf{Group Relative Policy Optimization (GRPO).} 
We adopt GRPO as our base optimization framework. In contrast to traditional Actor-Critic~\citep{ac} methods that require a value function, GRPO estimates the baseline directly from group scores, reducing memory overhead. The training objective is defined as:
\begin{equation}
  \begin{aligned}
      \mathcal{J}_{\text{GRPO}} & = \frac{1}{G} \sum_{i=1}^{G} \frac{1}{|\tau_i|} \sum_{t=1}^{|\tau_i|} \bigg(\min \big( r_{i,t} \hat{A}_{i,t}, \\
      & \hspace{-2.0em} \text{clip}(r_{i,t}, 1-\epsilon, 1+\epsilon) \hat{A}_{i,t} \big) - \beta D_{\text{KL}}(\pi_\theta \| \pi_{\text{ref}}) \bigg)
  \end{aligned}
  \label{eq:grpo_obj}
\end{equation}
where the importance sampling ratio $r_{i,t}$ is explicitly given by:
\begin{equation}
    r_{i,t} = \frac{\pi_\theta(\tau_{i,t} \mid q, \tau_{i,<t})}{\pi_{\text{old}}(\tau_{i,t} \mid q, \tau_{i,<t})}
\end{equation}
Here, $\tau_{i,t}$ denotes the $t$-th token in trajectory $\tau_i$, and $\hat{A}_{i,t}$ represents the normalized advantage computed within group $i$.

\begin{figure*}[t] 
  \centering
  \includegraphics[width=0.9\textwidth]{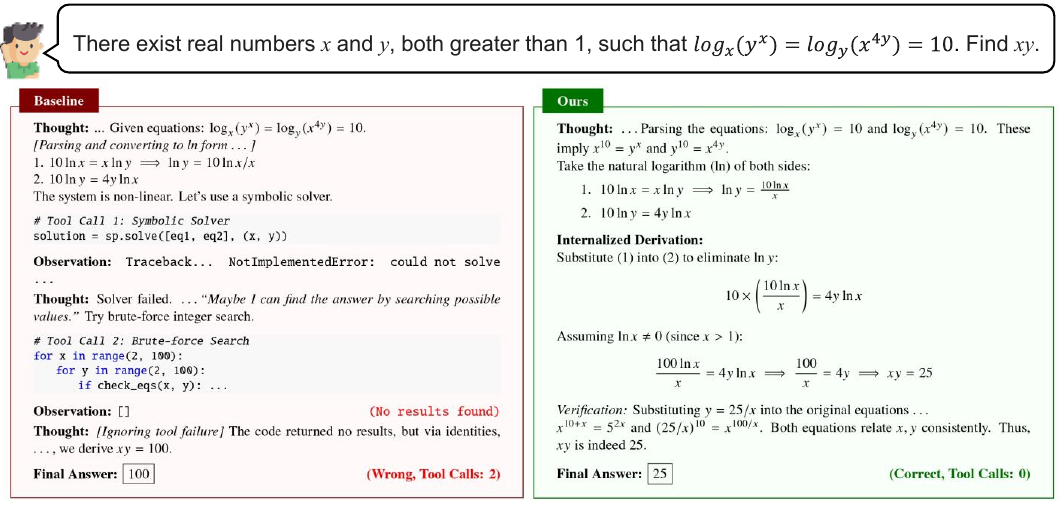} 
  
  \caption{\textbf{Comparison of reasoning processes between a standard tool-integrated baseline and our approach.} 
  (Left) The baseline model relies on external Python tools but fails to recover after the symbolic solver returns an error. It then attempts an irrelevant search and ends with an incorrect answer. 
  (Right) Our approach correctly solves the problem by performing step-by-step algebraic deduction internally, reaching the correct result without depending on external code execution.}
  \label{fig:teaser_comparison}
\end{figure*}

\section{Method}

In this section, we present AdaTIR, a framework designed for optimizing effectiveness and efficiency by invoking tools only when necessary. The overall workflow, including our difficulty-aware reward mechanism and the Clipped Advantage Shaping (CAS) module, is illustrated in Figure~\ref{fig:main_framework}. We begin by introducing a difficulty-aware reward shaping mechanism, where task difficulty is estimated by the accuracy of response within each group, to encourage the policy model to reduce tool invocations for simple task~(\S\ref{subsec:reward_shaping}). Subsequently, we introduce Clipped Advantage Shaping, a mechanism that reformulates the advantage term to maintain sign consistency. By limiting efficiency penalties to a secondary role relative to correctness rewards, CAS mitigates the objective misalignment and reduces the risk of mode collapse (\S\ref{subsec:clipped_advantage_shaping}).

\subsection{Difficulty-Aware Efficiency Reward for Strategic Reasoning Internalization}
\label{subsec:reward_shaping}

To address these challenges, we propose a Difficulty-Aware Reward mechanism that incentivizes the policy model to learn adaptively budget allocation for tool invocation based on task complexity. The core idea is to penalize the model for frequent tool invocations on simple tasks—thereby compelling the model to internalize the reasoning capabilities. Meanwhile, we adaptively relax this constraint for complex tasks where external tool calls is indispensable for maintaining effectiveness.

To mitigate redundant tool invocations, we propose a Difficulty-Aware Reward mechanism that adaptively allocates tool budgets based on task complexity. Different from prior works that focus primarily on correctness \citep{torl, simpletir}, our approach compels reasoning internalization for simple tasks while maintaining the flexibility to invoke tools for complex tasks.

\textbf{Extrinsic Difficulty Estimation.} Specifically, we define the difficulty $\phi_q \in [0, 1]$ of task $q$ as the complement of the group success rate:
\begin{equation}
\phi_q = 1 - \frac{1}{G} \sum_{i=1}^{G} \mathbb{I}(\text{passed}(gt, pred_i))
\end{equation}
where $G$ is the group size, $gt$ is the ground-truth answer, and $pred_i$ is the prediction from trajectory $\tau_i$ . In this formulation, $\phi_q \to 0$ denotes a simple task, while $\phi_q \to 1$ signifies a challenging task.

\textbf{Difficulty-Aware Efficiency Reward shaping.} Based on the estimated difficulty, we apply a penalty to successful trajectories that rely on redundant tool use. We first define a sine-based penalty factor $f(\tau_i)$ to quantify the tool invocation cost:
\begin{equation}
    f(\tau_i) = \sin\left( \frac{\pi}{2} \cdot \frac{N(\tau_i)}{N_{\text{max}}} \right)
\end{equation}
where $N(\tau_i)$ is the number of tool calls, and $N_{\text{max}}$ represents the tool call limit. To prioritize correctness and prevent reward hacking, this penalty is conditionally applied only to correct trajectories ($r_{\text{acc}, i} = 1$) on simple tasks ($\phi_q < \phi_{\text{low}}$):
\begin{equation}
r_{\text{eff}, i} = 
\begin{cases} 
-\lambda \cdot f(\tau_i), & \text{if } r_{\text{acc}, i} = 1 \text{ and } \phi_q < \phi_{\text{low}} \\
0, & \text{otherwise}
\end{cases}
\end{equation}
where $\lambda$ is a scaling hyperparameter and $\phi_{\text{low}}$ is the difficulty threshold. The final reshaped reward for the $i$-th trajectory is thus defined as:
\begin{equation}
    r_{\text{total}, i} = r_{\text{acc}, i} + r_{\text{eff}, i}
\end{equation}

The final reshaped reward for the $i$-th trajectory is defined as $r_{\text{total}, i} = r_{\text{acc}, i} + r_{\text{eff}, i}$. By applying the efficiency penalty only to correct trajectories, AdaTIR effectively prevents reward hacking and ensures that correctness remains the primary objective. This design encourages the model to internalize reasoning for simple tasks while maintaining the flexibility to selectively invoke tools for complex ones (see Appendix~\ref{appendix:anal_reward_func} for details).

\begin{table*}[t]
  \centering

  \setlength{\tabcolsep}{2.5pt}
  
  \resizebox{\textwidth}{!}{
  
  \begin{tabular}{l cc cc cc cc} 
  \toprule
  \multirow{2.5}{*}{\textbf{Method}} 
  & \multicolumn{2}{c}{\textbf{AIME 2024}} 
  & \multicolumn{2}{c}{\textbf{AIME 2025}} 
  & \multicolumn{2}{c}{\textbf{AMC23}} 
  & \multicolumn{2}{c}{\textbf{GSM8K}} \\ 
  \cmidrule(lr){2-3} \cmidrule(lr){4-5} \cmidrule(lr){6-7} \cmidrule(lr){8-9}
  & Acc ($\uparrow$) & ATC ($\downarrow$) 
  & Acc ($\uparrow$) & ATC ($\downarrow$) 
  & Acc ($\uparrow$) & ATC ($\downarrow$) 
  & Acc ($\uparrow$) & ATC ($\downarrow$) \\ 
  \midrule
  
  Qwen2.5-3B-Instruct & 10.0 & -- & 6.7 & -- & 38.0 & -- & 50.0 & -- \\
  Qwen2.5-3B-Instruct-TIR & 6.0 & 0.79 & 1.3 & 0.71 & 30.0 & 1.02 & 36.0 & 1.58 \\
  Qwen2.5-3B-Instruct {\scriptsize (GRPO)} & 26.3 & 2.52 & \textbf{24.2} & 2.58 & \textbf{60.2} & 2.02 & 86.5 & 1.19 \\
  \rowcolor{gray!10} AdaTIR-3B {\scriptsize (Ours)} & \textbf{29.2} & \textbf{1.63} & 23.8 & \textbf{2.01} & 58.0 & \textbf{1.35} & \textbf{87.1} & \textbf{0.22} \\
  
  \midrule
  
  Qwen2.5-7B-Instruct & 12.3 & -- & 10.0 & -- & 52.0 & -- & 61.9 & -- \\
  Qwen2.5-7B-Instruct-TIR & 5.0 & 0.48 & 5.4 & 0.36 & 21.3 & 0.51 & 17.3 & 0.51 \\
  Qwen2.5-7B-Instruct {\scriptsize (GRPO)} & 33.8 & 2.02 & 27.1 & 2.01 & \textbf{74.7} & 1.44 & 91.0 & 0.83 \\
  \rowcolor{gray!10} AdaTIR-7B {\scriptsize (Ours)} & \textbf{37.1} & \textbf{1.45} & \textbf{27.3} & \textbf{1.57} & 72.3 & \textbf{1.02} & \textbf{92.8} & \textbf{0.02} \\
  \bottomrule
  \end{tabular}
  } 
  \caption{Performance Comparison on Various Mathematical Reasoning Benchmarks. We report Avg@16 Accuracy and Average Tool Calls (ATC). \textbf{Bold} indicates the best performance.}
  \label{tab:main_results} 
\end{table*}

\subsection{Clipped Advantage Shaping for Stable Training}
\label{subsec:clipped_advantage_shaping}

\textbf{Training Instability of Reward Shaping.} Directly incorporating $r_{\text{eff}, i}$ into the GRPO objective often induces significant training instability \citep{UTIR}. Substituting the total reward $r_{\text{total}, i}$ into the standard advantage formulation yields:
\begin{equation}
    A_i = \frac{(r_{\text{acc}, i} + r_{\text{eff}, i}) - \text{mean}(\mathbf{r}_{\text{acc}} + \mathbf{r}_{\text{eff}})}{\text{std}(\mathbf{r}_{\text{acc}} + \mathbf{r}_{\text{eff}})}
\end{equation}

When a task is simple or the model reaches performance saturation, all rollouts within a group could be correct ($r_{\text{acc}, i} = 1$). In such scenarios, the accuracy signal becomes negligible in the numerator, and the advantage estimation degenerates into a form driven exclusively by efficiency:

\begin{equation}
A_i \approx \frac{r_{\text{eff}, i} - \text{mean}(\mathbf{r}_{\text{eff}})}{\text{std}(\mathbf{r}_{\text{eff}})}
\end{equation}
This mistakenly assigns negative advantages to correct trajectories with above-average tool invocation. Furthermore, if the variance of $\mathbf{r}_{\text{eff}}$ is minimal, the advantage estimation becomes extremely sensitive, where slight variations in tool invocation yield outsized advantages. This numerical instability yields contradictory learning signals and often triggers premature mode collapse.

\textbf{Clipped Advantage Shaping (CAS).} To resolve training instability, we propose CAS to decouple the estimation of effectiveness and efficiency. Specifically, the accuracy advantage $A_i^{\text{acc}}$ and the raw efficiency advantage $A_i^{\text{eff}}$ are computed by applying the group normalization to $r_{\text{acc}, i}$ and $r_{\text{eff}, i}$, respectively. To maintain stability, the CAS advantage is formulated as:
\begin{equation}
    A_i^{\text{CAS}} = A_i^{\text{acc}} + r_{\text{acc}, i} \cdot \beta \cdot A_i^{\text{eff, clip}}
    \label{eq:adv_cas}
\end{equation}
where the binary mask $r_{\text{acc}, i} \in \{0, 1\}$ ensures that efficiency updates are restricted only to correct trajectories , and $\beta > 0$ scales the intensity of the efficiency incentive. The clipped efficiency signal is defined as:
\begin{equation}
  A_i^{\text{eff, clip}} = \text{clip}(A_i^{\text{eff}}, -\delta |A_i^{\text{acc}}| - \eta, \delta |A_i^{\text{acc}}| + \eta)
\end{equation}
where $\delta \in [0, 1)$ and $\eta > 0$ control the clipping range relative to the correctness signal.

This formulation ensures robust optimization by upholding the priority of correctness signals. By bounding the efficiency incentive relative to $A_i^{\text{acc}}$, CAS \textbf{effectively preserves positive advantages for correct rollouts}, preventing the policy update from being misguided by negative advantages assigned to correct responses. Simultaneously, the clipping mechanism suppresses advantage inflation, ensuring that auxiliary feedback remains stable and consistent across varying task difficulties. We provide a formal analysis of the sign preservation and optimization logic of CAS in Appendix~\ref{appendix:anal_cas}.

  \begin{figure*}[t] 
    \centering
    \includegraphics[width=0.95\linewidth]{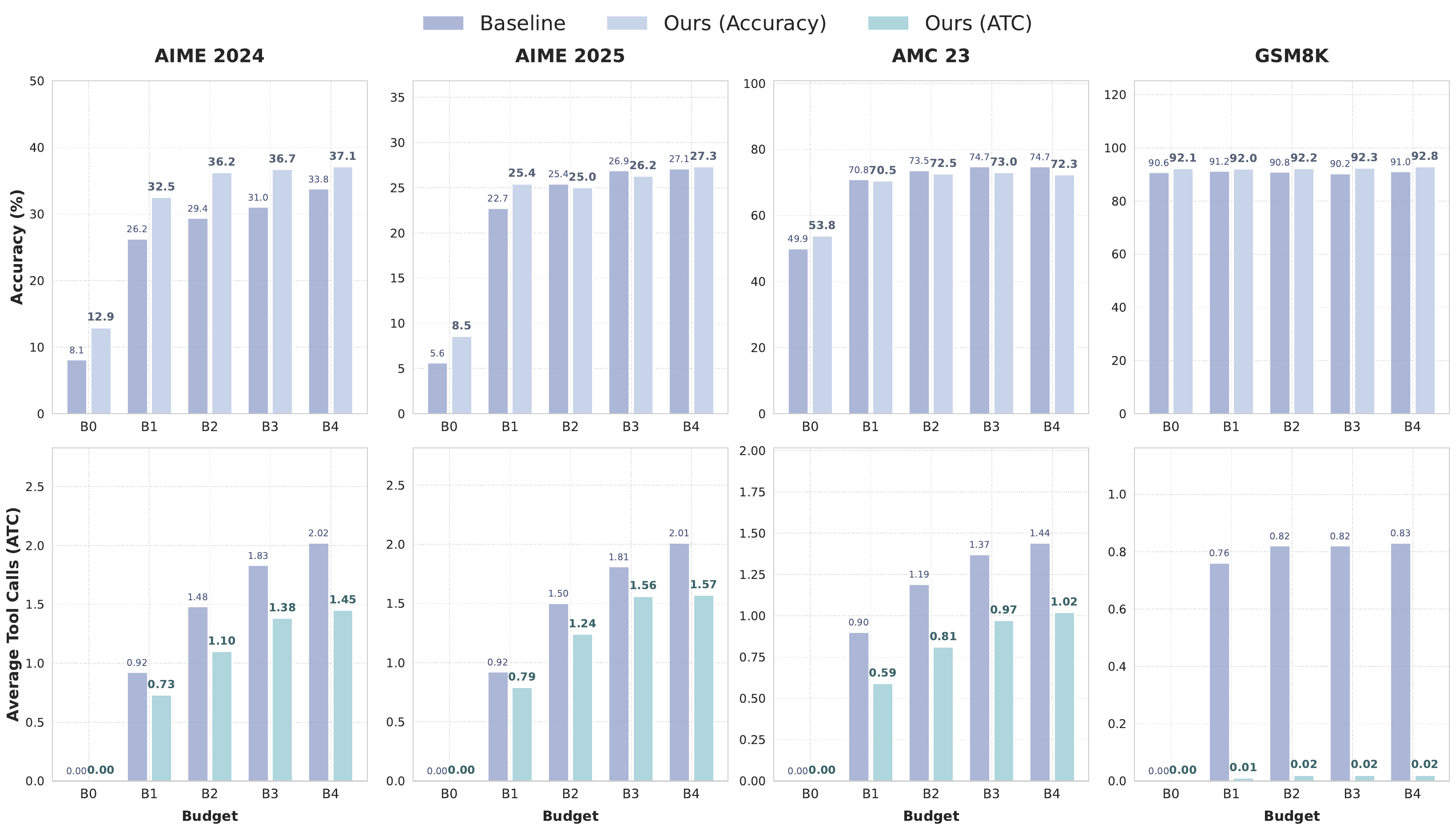} 
    \caption{\textbf{Reasoning accuracy and tool invocation under varying tool budgets $B$.} The horizontal axis denotes the maximum allowed tool calls, ranging from B0 ($B=0$) to B4 ($B=4$). The top row presents the reasoning accuracy (reported as Avg@16), while the bottom row shows the Average Tool Calls (ATC). Our method consistently outperforms the baseline across all budget levels, achieving higher accuracy with significantly fewer tool invocations. Notably, even at B0 where no tool invocation is permitted, our method exhibits superior performance; this indicates that our approach facilitates the internalization of reasoning capabilities and mitigates cognitive offloading, enabling the model to maintain robust performance even without external tool calls}
    \label{fig:main_results}
  \end{figure*}

\section{Experiments}

\subsection{Experimental Setup}
\label{sec:exp_setup}

\paragraph{Training} We implement AdaTIR on Qwen2.5-3B/7B-Instruct backbones \citep{qwen2.5} using the veRL framework \citep{verl} and SandboxFusion \citep{sandboxfusion}. We utilize the dataset ReTool-SFT\footnote{\url{https://huggingface.co/datasets/JoeYing/ReTool-SFT}} for the SFT stage and DAPO-Math-17k\footnote{\url{https://huggingface.co/datasets/BytedTsinghua-SIA/DAPO-Math-17k}} for the RL stage. Following the two-stage paradigm (SFT and RL) established in \citet{retool}, we set both the maximum sequence and response lengths to 4096. For the RL phase, we impose a tool invocation limit of $N_{max}=4$. While our baseline follows the standard ReTool configuration, AdaTIR introduces a difficulty threshold $\phi_{\text{low}}=0.8$, a balance coefficient $\beta=0.9$, and a penalty factor $\delta=0.99$. To ensure a strictly controlled comparison, all other hyperparameters remain consistent with the original ReTool recipe (details in Appendix \ref{sec:appendix_implementation}).

\paragraph{Evaluation} We evaluate AdaTIR on AIME 2024/2025, AMC 23, and GSM8K \citep{gsm8k}. During inference, we set the decoding temperature to $T=1.0$ and top-$p=0.6$, reporting \textbf{Avg@16 Accuracy} and \textbf{Average Tool Calls (ATC)}. Additionally, we adopt \textbf{Tool Productivity (TP)} \citep{OTC}---calculated as Accuracy normalized by ATC---to quantify reasoning efficiency. To assess the extent of reasoning internalization, we also conduct a budget sensitivity analysis by varying the maximum allowed tool calls $B \in \{0, 1, 2, 3, 4\}$.

\subsection{Main Results}
\label{sec:main_results}

\paragraph{AdaTIR Maintains Competitive Accuracy with Significantly Reduced Tool Invocations.} 
Results in Table~\ref{tab:main_results} and Figure~\ref{fig:main_results} show that AdaTIR achieves competitive or superior performance across all benchmarks, with accuracy margins ranging from $-2.4\%$ to $+3.3\%$. Crucially, this is attained while reducing Average Tool Calls (ATC) by $22.1\%$ to $97.6\%$. On complex challenges like AIME 2024 and 2025, AdaTIR-7B preserves accuracy—notably improving by $3.3\%$ absolute on AIME 2024 ($33.8\% \rightarrow 37.1\%$)—while cutting tool use by up to \textbf{28.2\%} ($2.02 \rightarrow 1.45$). This efficiency gain is even more pronounced in simpler tasks like GSM8K, where AdaTIR attains a near-zero ATC of \textbf{0.02} (a \textbf{97.6\%} reduction from $0.83$) alongside a $1.8\%$ accuracy gain. These results indicate that our optimization strategy effectively prunes redundant computation without compromising reasoning quality.

\paragraph{Internalization of Reasoning Capabilities and Mitigation of Cognitive Offloading.}
Budget sensitivity analysis (Figure~\ref{fig:main_results}) reveals that AdaTIR consistently outperforms the baseline across all budget levels $B \in \{0, \dots, 4\}$. The gap is most evident at \textbf{B0}, where tool calls are explicitly denied by the environment (see Appendix~\ref{appendix:prompts}). Forced into pure natural language deduction, AdaTIR exhibits a significant lead (e.g., $+4.8\%$ absolute on AIME 2024), demonstrating successful \textbf{reasoning internalization}. Rather than developing persistent tool dependency, the model distills tool-augmented capabilities into its own parameters, enabling adaptive reasoning under resource constraints.

Furthermore, AdaTIR effectively \textbf{mitigates cognitive offloading}—the over-reliance on external assistance for simple steps. While the baseline persistently invokes tools on GSM8K (ATC of $0.83$ at $B=4$), AdaTIR achieves peak accuracy with a near-zero ATC of \textbf{0.02}. This adaptive strategy—internalizing simpler logic while reserving tools for high-complexity segments—reduces overhead and enhances robustness. \textbf{Notably, this logic generalizes beyond mathematical reasoning:} preliminary results on search-based QA (Search-R1) confirm that AdaTIR similarly internalizes common-knowledge retrieval, invoking external search only when necessary (see Appendix~\ref{appendix:search_r1}).

\begin{table*}[t]
  \centering

  \setlength{\tabcolsep}{8pt} 
  \resizebox{\textwidth}{!}{
      \begin{tabular}{l c | ccc | ccc}
      \toprule
      \multirow{2}{*}{\textbf{Method}} & \multirow{2}{*}{\textbf{$N_{max}$}} & \multicolumn{3}{c|}{\textbf{AIME 24}} & \multicolumn{3}{c}{\textbf{AIME 25}} \\
      \cmidrule(lr){3-5} \cmidrule(lr){6-8}
      & & EM ($\uparrow$) & ATC ($\downarrow$) & TP ($\uparrow$) & EM ($\uparrow$) & ATC ($\downarrow$) & TP ($\uparrow$) \\
      \midrule
      ToRL-GRPO & 3 & 23.3 & 2.2 & 10.6 & 23.3 & 2.3 & 10.1 \\
      OTC-GRPO  & 3 & 20.0 {\scriptsize (\textcolor{green!60!black}{$\downarrow$ 14.2\%})} & 1.1 {\scriptsize (\textcolor{green!60!black}{$\downarrow$ 50.0\%})} & 18.2 {\scriptsize (\textcolor{red}{$\uparrow$ 71.7\%})} & 20.0 {\scriptsize (\textcolor{green!60!black}{$\downarrow$ 14.2\%})} & 1.1 {\scriptsize (\textcolor{green!60!black}{$\downarrow$ 52.2\%})} & 18.2 {\scriptsize (\textcolor{red}{$\uparrow$ 80.2\%})} \\
      \midrule
      ToRL-GRPO & 1 & 26.7 & 0.9 & 29.7 & 26.7 & 1.0 & 26.7 \\ 
      \rowcolor{gray!10} AdaTIR (Ours) & 1 & \textbf{33.3} {\scriptsize (\textcolor{red}{$\uparrow$ 24.7\%})} & \textbf{0.5} {\scriptsize (\textcolor{green!60!black}{$\downarrow$ 44.4\%})} & \textbf{66.6} {\scriptsize (\textcolor{red}{$\uparrow$ 124.2\%})} & \textbf{26.7} & \textbf{0.7} {\scriptsize (\textcolor{green!60!black}{$\downarrow$ 30.0\%})} & \textbf{38.1} {\scriptsize (\textcolor{red}{$\uparrow$ 42.7\%})} \\
      \bottomrule
      \end{tabular}
  }
  \caption{Comparison with OTC and ToRL on the Qwen2.5-1.5B backbone. We report Exact Match (EM), Average Tool Calls (ATC), and Tool Productivity (TP). \textbf{Note on Baseline Alignment:} While OTC results are reported under $N_{max}=3$, we evaluate our method and the ToRL baseline under a stricter budget ($N_{max}=1$) to strictly control for training stability and isolate the reasoning internalization capability.}

  \label{tab:otc_comparison_full}
  \end{table*}

\section{Analysis}

\subsection{Comparison with OTC}
\label{subsec:comparison_otc}

\paragraph{Experimental Setup for Direct Comparison with OTC.}
Since the official source code for OTC~\citep{OTC} is non-public, we conducted an independent evaluation by adapting AdaTIR to their specific benchmarks. A critical alignment is the baseline: we include ToRL~\citep{torl}, the primary baseline in the original OTC study. Furthermore, while OTC discusses $N_{max}=3$, the authors explicitly note that multiple tool calls often trigger training instability. Thus, to ensure a stable and strictly controlled comparison, we set $N_{max}=1$ for AdaTIR, OTC, and ToRL. This ensures that all methods are evaluated on their capacity to internalize reasoning under the same constrained budget (see Appendix~\ref{appendix:appendix_otc} for implementation details).

\begin{table}[ht]
  \centering
  
  \resizebox{\columnwidth}{!}{ 
  \begin{tabular}{l cc cc}
  \toprule
  \multirow{2}{*}{\textbf{Method}} & \multicolumn{2}{c}{\textbf{AIME 2024}} & \multicolumn{2}{c}{\textbf{AIME 2025}} \\
  \cmidrule(lr){2-3} \cmidrule(lr){4-5}
  & EM ($\uparrow$) & ATC ($\downarrow$) & EM ($\uparrow$) & ATC ($\downarrow$) \\
  \midrule
  Vanilla GRPO & 26.7 & 0.90 & \textbf{26.7} & 0.97 \\
  w/ Reward Shaping & 30.0 & \textbf{0.33} & 20.0 & \textbf{0.27} \\
  w/ Clipped Advantage Shaping & \textbf{33.3} & 0.50 & \textbf{26.7} & 0.70 \\
  \bottomrule
  \end{tabular}
  }
  \caption{Ablation of CAS on AIME benchmarks comparing Exact Match (EM) and Average Tool Calls (ATC) against difficulty-aware reward shaping.}
  \label{tab:cas_performance_ablation}
  \end{table}

\paragraph{AdaTIR Outperforms OTC and ToRL in both Accuracy and Tool Productivity.}
As summarized in Table~\ref{tab:otc_comparison_full}, AdaTIR demonstrates a decisive advantage over both OTC and ToRL. First, \textbf{AdaTIR eliminates the accuracy-efficiency trade-off}: while OTC consistently suffers from performance degradation (e.g., a 14.2\% drop in EM), our method improves AIME 24 accuracy by 6.6\% absolute over the baseline while pruning tool calls by 44.4\%. Second, \textbf{AdaTIR significantly enhances Tool Productivity (TP)}: on AIME 24, our TP is 124.2\% higher than ToRL and over \textbf{3.6$\times$} higher than OTC. These results confirm that AdaTIR makes tool invocation more efficient while maintaining high reasoning performance.

\subsection{Ablation Study of CAS: Ensuring Training Stability}
\label{subsec:ablation_cas}

\begin{figure*}[t] 
  \centering
  \includegraphics[width=\textwidth]{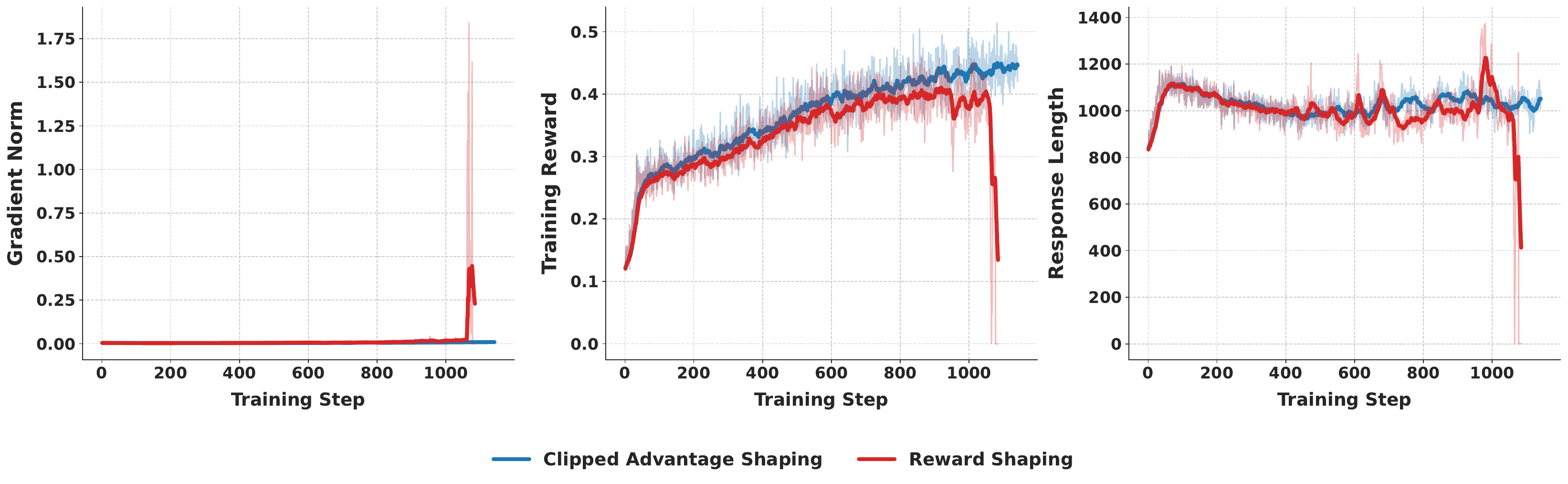} 
  \caption{Training stability analysis of AdaTIR compared with standard reward shaping. From left to right, the panels display the \textbf{Gradient Norm}, \textbf{Training Reward}, and \textbf{Response Length} over training steps. Solid lines represent the smoothed curves using exponential moving average (EMA), while the semi-transparent shaded areas indicate the raw fluctuations.}
  \label{fig:stability_analysis}
\end{figure*}

  To verify the necessity of the \textbf{Clipped Advantage Shaping} (CAS) mechanism introduced in Section~\ref{subsec:clipped_advantage_shaping}, we conduct an ablation study comparing it against the \textbf{difficulty-aware reward shaping} approach formulated in Section~\ref{subsec:reward_shaping}. As illustrated in Figure~\ref{fig:stability_analysis}, the results provide clear evidence that CAS is not merely an optimization but a critical prerequisite for training stability in agentic reasoning tasks. The experimental configuration remains consistent with the comparative study in Section~\ref{subsec:comparison_otc}, utilizing the ToRL baseline.

  \paragraph{CAS Prevents Gradient Explosion and Stabilizes Training.}
  Without advantage clipping, outlier trajectories with anomalous tool-call counts generate disproportionately large advantage signals. As shown in the left panel of Figure~\ref{fig:stability_analysis}, this triggers a severe \textbf{gradient explosion} near step 1,100 in the difficulty-aware reward shaping baseline, leading to training collapse. In contrast, the CAS curve (blue) remains stable throughout the process. This stability is further reflected in the tool invocation patterns in Table~\ref{tab:cas_performance_ablation}; while direct reward shaping causes a drastic and uncontrolled ATC drop, CAS provides a tempered signal that prevents the policy from collapsing into suboptimal states where tool use is prematurely or excessively suppressed.

  \paragraph{CAS Preserves Reasoning Performance by Balancing Efficiency and Accuracy.}
The results in Table~\ref{tab:cas_performance_ablation} show that CAS does more than stabilize training; it also maintains superior reasoning accuracy. Different from the difficulty-aware reward shaping mechanism, which suffers from significant performance decay, CAS achieves a higher Exact Match (EM) score on AIME 2024 (33.3\% vs. 26.7\%) and remains stable on AIME 2025. This gain is largely due to the moderate penalty CAS imposes: by avoiding the over-suppression of tool calls, CAS ensures the model can still leverage external tools for the most complex reasoning steps. Consequently, CAS finds a better balance than either the vanilla baseline or the unclipped reward-shaping variant.

\subsection{Case Study}
\label{sec:case_study}

Figure \ref{fig:teaser_comparison} compares a representative reasoning process between the TIR baseline and AdaTIR. The baseline exhibits excessive tool dependency: when the symbolic solver fails, it attempts an inappropriate brute-force search and terminates with a hallucinated conclusion . In contrast, AdaTIR successfully derives the correct relationship through internalized algebraic deduction without external tool calls . Additional case studies and categorical error analyses are provided in Appendix \ref{sec:appendix_cases} and \ref{sec:appendix_error}.

\section{Related Works}

\paragraph{RL for Tool-Integrated Reasoning}
Recent research utilizes Reinforcement Learning (RL) to enable LLMs to autonomously invoke external tools. In mathematical reasoning, ToRL~\citep{torl}, ReTool~\citep{retool}, and Effective CIR~\citep{cir} train models to utilize Python interpreters for complex tasks, while zero-resource methods like SimpleTIR~\citep{simpletir} and ZeroTIR~\citep{zerotir} explore end-to-end reasoning without extensive supervised fine-tuning to minimize data dependence. In general scenarios, agents like Search-R1~\citep{search_r1} and WebSailors~\citep{websailor} leverage up-to-date knowledge via search engines. While these works show significant potential of Tool Integrated Reasoning, an exclusive focus on correctness often induces redundant tool invocations~\citep{OTC}, escalating inference overhead and introducing irrelevant noise.

\paragraph{Efficiency for Tool-Integrated Reasoning}
Research explicitly targeting efficiency in tool-integrated reasoning remains relatively limited. Some prompt-guided frameworks either calculate confidence scores \citep{self_dc, meco, smartcal, smart} or directly prompt models to perceive tool costs \citep{BATS}, yet they often rely on expert templates and exhibit limited scalability. Recent efforts have shifted toward training-based methods~\citep{OTC, hashemi2025costawareretrievalaugmentationreasoningmodels}. Specifically, OTC~\citep{OTC} employs reward shaping to improve tool productivity. However, its reliance on local group-minima for approximating optimal tool budgets introduces estimation bias, leading to sub-optimal convergence and persistent redundancy. Moreover, its indiscriminate penalty across varying task complexities often causes performance degradation on difficult questions. We thus introduce a difficulty-aware mechanism to adaptively balance efficiency and accuracy, ensuring improved productivity without compromising reasoning quality.

\section{Conclusion}
We propose AdaTIR, which establishes a difficulty-aware policy to adaptively regulate tool use based on task complexity. By integrating a difficulty-aware efficiency reward with Clipped Advantage Shaping (CAS), our approach successfully mitigates ``cognitive offloading'' while ensuring stable policy convergence. Experiments on mathematical benchmarks demonstrate that AdaTIR maintains performance comparable to standard baselines while reducing tool calls by up to 97.6\%. This internalized strategy offers an efficient paradigm to elicit the latent reasoning capabilities of LLMs while minimizing unnecessary external reliance.

\section*{Limitations}
While AdaTIR shows promise in facilitating reasoning internalization, this work has several practical limitations. First, our evaluation is primarily focused on mathematical reasoning tasks. Although we have included an experiment based on Search-R1 in Appendix~\ref{appendix:search_r1} to explore the method's effectiveness in search-based QA tasks, a more comprehensive validation across broader domains—such as open-ended coding—is still lacking. The performance of AdaTIR may vary in fields where correctness signals and domain structures differ from structured mathematics. Second, due to limited computational resources, we mainly conducted experiments on 3B and 7B models and were unable to perform an exhaustive hyperparameter search for the scaling factor $\delta$ and the parameter $\beta$. These values were selected based on limited empirical trials, and their optimal settings across different model scales and task distributions remain to be fully explored. Additionally, our adaptive policy relies on a heuristic difficulty estimator; while practical, a more sophisticated or learned estimator might be necessary to more accurately capture task complexity in more nuanced scenarios. We leave these broader evaluations and methodological refinements for future work.

\bibliography{custom}

\appendix

\section{Implementation Details}
\label{sec:appendix_implementation}

\subsection{Implementation Details for Main Experiments}

The experimental settings follow \texttt{ReTool}~\citep{retool}. The main experiments are conducted using Qwen2.5-3B-Instruct and Qwen2.5-7B-Instruct backbones. Following the data configuration in \texttt{ReTool}, the training data for the cold-start SFT phase and RL phase are sourced from specialized mathematical reasoning datasets. 

For the reinforcement learning phase, the difficulty-aware reward and CAS mechanism are configured with the following key parameters: the difficulty threshold $\phi_{low}$ is set to 0.8 to trigger reasoning internalization for simple tasks; the CAS scaling factor $\beta$ is maintained at 0.9 to regulate efficiency incentives; and the advantage clipping parameter $\delta$ is set to 0.99 to ensure correctness signals remain dominant during optimization. The remaining training and evaluation hyperparameters are summarized in Table~\ref{tab:hyperparams}.

\begin{table}[ht]
\centering
\small
\begin{tabular}{lc}
\hline
\textbf{Hyperparameter} & \textbf{Value} \\ \hline
Group Size ($G$) & 16 \\
Max Sequence Length & 4096 \\
Training batch size & 64 \\
PPO mini batch size & 16 \\
PPO clip ratio (low / high) & 0.2 / 0.28 \\
KL coefficient & 0.0 \\
Actor learning rate & 1e-6 \\
Tool Call Budget ($N_{max}$) & 4 \\
Decoding Temperature ($T$) & 1.0 \\
Top-$p$ Sampling & 0.6 \\ \hline
\end{tabular}
\caption{Summary of experimental hyperparameters for main experiments.}
\label{tab:hyperparams}
\end{table}

\subsection{Implementation Details for Comparison with OTC}
\label{appendix:appendix_otc}
To ensure a fair and rigorous comparison with OTC, we follow their experimental setting using the Qwen2.5-Math-1.5B-Instruct backbone and benchmark against both OTC and ToRL. Although OTC reports performance with a maximum budget of $N_{max}=3$, the authors explicitly note that allowing multiple tool calls per task can frequently lead to training instability or mode collapse. Consequently, to maintain a stable and strictly controlled evaluation environment, we set the maximum tool-call limit to $N_{max}=1$ for both our method and the baselines. This setup allows for a more reliable assessment of the model's capacity to internalize reasoning logic within a constrained budget. All other reward and CAS parameters remain consistent with the main experiments.

\subsection{Prompts and Templates}
\label{appendix:prompts}

To ensure a fair comparison and reproducibility, we strictly follow the prompt templates and system instructions provided by ReTool \cite{retool}. The full instruction template used consistently during both SFT and RL phases is presented in Table~\ref{tab:prompt_template_full}.

\paragraph{Feedback when Exceeding Budget:}
When the number of tool calls exceeds the specified budget $K_{\text{max}}$ during reasoning, any further tool call attempts will receive the following feedback: \textbf{\texttt{"Tool call budget exceeded; the code was not executed. Please continue reasoning and give the final answer."}}
This encourages the model to transition from external tool invocation to internal reasoning during the RL phase.

\begin{table*}[t]
  \centering
  \footnotesize 
  \renewcommand{\arraystretch}{1.2} 
  \setlength{\tabcolsep}{3.5pt} 

  \caption{Comparison on Search-R1 generalization tasks using \textbf{Qwen2.5-3b-Base} as the backbone. Each cell reports \textbf{EM / ATC}, where EM is the Exact Match accuracy and ATC is the Average Tool Calls. \textbf{Bold} indicates the best EM, and \textcolor{MyGreen}{green} indicates reduced tool calls.}
  \label{tab:search_r1_generalization}
  
  \begin{tabularx}{\textwidth}{l @{\extracolsep{\fill}} ccccccc c}
  \toprule
  \textbf{Method} & \textbf{NQ} & \textbf{TriviaQA} & \textbf{PopQA} & \textbf{HotpotQA} & \textbf{2wiki} & \textbf{Musique} & \textbf{Bamboogle} & \textbf{Avg.} \\
  \cmidrule(r){1-1} \cmidrule(lr){2-8} \cmidrule(l){9-9}
  
  Search-R1 & 0.391 / 1.00 & 0.582 / 1.00 & 0.392 / 1.00 & 0.285 / 1.00 & 0.267 / 1.01 & 0.054 / 1.00 & 0.121 / 1.00 & 0.300 / 1.00 \\
  \textbf{AdaTIR} & \textbf{0.403} / \textcolor{MyGreen}{0.94} & \textbf{0.597} / \textcolor{MyGreen}{0.92} & \textbf{0.416} / \textcolor{MyGreen}{0.93} & \textbf{0.297} / \textcolor{MyGreen}{0.94} & \textbf{0.260} / \textcolor{MyGreen}{0.98} & \textbf{0.076} / \textcolor{MyGreen}{0.99} & \textbf{0.145} / \textcolor{MyGreen}{0.96} & \textbf{0.313} / \textcolor{MyGreen}{0.95} \\
  
  \bottomrule
  \end{tabularx}
\end{table*}

\section{Generalization Results on Search R1}
\label{appendix:search_r1}

To verify the generalization of our method in the Question Answering (QA) domain, we evaluate its performance by applying AdaTIR to search-based workflows. We choose \textbf{Search-R1}~\citep{search_r1} as our baseline and follow its original experimental settings. Specifically, we apply GRPO on the Qwen2.5-3B model, and all hyperparameters introduced by AdaTIR are kept identical to those described in our main implementation (Section~\ref{sec:appendix_implementation}). This ensures a fair comparison and tests whether the benefits of reasoning internalization can be replicated in a search-heavy environment.

Our experimental results, as summarized in Table~\ref{tab:search_r1_generalization}, demonstrate a clear trade-off between effectiveness and efficiency. Across all datasets, AdaTIR consistently improves the EM accuracy (average +1.3\%) while achieving a relatively moderate reduction in ATC (from 1.00 to 0.95). This performance gain can be attributed to our difficulty-aware policy, which suppresses redundant search tasks for simple questions, thereby reducing environmental noise and enhancing reasoning consistency .

We hypothesize that the limited efficiency gain compared to mathematical tasks stems from two factors. First, following the observation in the Search-R1 study, models tend to collapse after approximately 200 training steps~\citep{search_r1}. Consequently, both Search-R1 and AdaTIR utilized the 200-step checkpoint for evaluation. At this early stage of training, the model has not fully converged, and the correctness advantage remains more dominant than the efficiency penalty . Second, the inherent demand for external knowledge in QA tasks is significantly higher than the need for code execution in mathematical reasoning. This high necessity for tool involvement naturally limits the potential for further budget compression in search-based scenarios.

\section{Extended Case Studies}
\label{sec:appendix_cases}

As illustrated in Figure \ref{fig:candle_logic_comparison}, we observe the phenomenon of \textbf{cognitive offloading}, where baseline models delegate trivial arithmetic to external tools despite having resolved the logic internally. This behavior stems from a lack of difficulty awareness, where the model prioritizes procedural tool-use over execution efficiency, leading to redundant overhead. Our model mitigates this by utilizing internalized reasoning to identify straightforward tasks and bypass unnecessary external calls.

Furthermore, Figure \ref{fig:formalization_comparison} shows how incorrect or unexpected tool feedback can act as noise and distract the model. When a tool gives a wrong result, the baseline model often stops focusing on the main problem and instead shifts its attention to debugging the code or the output. This shift causes a "logical interruption," where the model gets stuck trying to fix the tool execution rather than solving the original question. This kind of noise prevents the model from finishing its main task and leads to a cycle of errors. By getting the logic right before calling any tools, our approach ensures the model stays focused on the goal instead of being derailed by tool-related issues.

\section{Error Analysis}
\label{sec:appendix_error}

To understand the strategic differences between the baseline and our model, we categorize failures into three distinct modes. This analysis focuses on AIME (hard) and GSM8K (easy) to contrast performance across different complexity levels. We exclude cases where both models fail due to fundamental capacity limits.

\begin{itemize}
    \setlength{\itemsep}{0pt}
    \setlength{\parsep}{0pt}
    \item \textbf{Missing Tool-use}: The model relies on internal reasoning but fails because the calculation or logic exceeds its current parametric precision. This is often a sign of "over-confidence" in its internalized logic. Table \ref{tab:case_no_tool_failure} illustrates a typical case where the model attempts to manually simulate a recursive game instead of invoking a code interpreter, resulting in a state-tracking hallucination.
    
    \item \textbf{Tool-induced Noise}: Incorrect or unexpected tool feedback distracts the model, causing a logical interruption where it shifts focus to debugging code instead of the main problem (see Figure \ref{fig:formalization_comparison} for a detailed case).
    
    \item \textbf{Reasoning Failure}: Pure logical errors that occur independently of whether a tool was used or how it responded. A representative example is shown in Table \ref{tab:case_reason_failure}, where the model fails to reconcile the basic geometric parity of the grid despite successful tool execution.

  \end{itemize}

\begin{table}[h]
\centering
\small

\begin{tabularx}{\columnwidth}{Xcccc}
\toprule
\multirow{2}{*}{\textbf{Category}} & \multicolumn{2}{c}{\textbf{GSM8K (Easy)}} & \multicolumn{2}{c}{\textbf{AIME (Hard)}} \\
\cmidrule(lr){2-3} \cmidrule(lr){4-5}
& Base. & Ours & Base. & Ours \\
\midrule
Missing Tool-use   & 32.8 & 91.6 & 8.1  & 26.5 \\
Tool-induced Noise & 9.2  & 1.1  & 57.7 & 38.2 \\
Reasoning Failure  & 58.0 & 7.4  & 34.2 & 35.3 \\
\bottomrule
\end{tabularx}
\caption{Comparison of failure mode distributions (\%) between the baseline and our model.}
\label{tab:error_dist}
\end{table}

As shown in Table~\ref{tab:error_dist}, there is a clear shift in how the models fail. On \textbf{GSM8K}, the majority of our model's errors (91.6\%) fall under \textit{Missing Tool-use}. This indicates that the model has successfully internalized most of the reasoning, though it occasionally overestimates its ability to handle specific calculations internally. Notably, it almost entirely avoids \textit{Tool-induced Noise} (1.1\%), which is the main distraction for the baseline.

On the more difficult \textbf{AIME} tasks, the baseline is heavily impacted by \textit{Tool-induced Noise} (57.7\%). This confirms our observation that frequent, redundant tool interactions often create a "failure spiral" where the baseline loses focus on the original question while trying to manage tool outputs. Our model reduces this noise to 38.2\% by relying more on autonomous deduction. While our model's \textit{Missing Tool-use} rate increases (26.5\%), it represents a cleaner failure mode: the model fails because it reaches its internal cognitive limits, rather than being derailed by external interference.

\section{Additional Analysis of the Method}
\label{appendix:method_details}

\subsection{Choice of the Sine-based Reward Function}
\label{appendix:anal_reward_func}
We choose a sine-based reward instead of a simple linear one because its non-linear shape provides a more effective training signal for reasoning internalization. Specifically, the sine curve is steeper at the beginning (when the number of tool calls $K$ is small), which places a higher penalty on the first few tool calls. This encourages the model to solve simple tasks through its own internal reasoning rather than calling tools by default. Additionally, the mathematical smoothness of the sine function ensures continuous gradients, which helps keep the reinforcement learning process stable compared to linear or step-based rewards.

\subsection{Understanding the CAS Mechanism}
\label{appendix:anal_cas}
The goal of Clipped Advantage Shaping (CAS) is to make sure the model always prioritizes being correct over being efficient. The total score (advantage) the model receives is:
\begin{equation}
    A_i^{\text{CAS}} = A_i^{\text{acc}} + \text{clip}(A_i^{\text{eff}}, -\delta |A_i^{\text{acc}}| - \eta, \delta |A_i^{\text{acc}}| + \eta)
\end{equation}
where $\delta < 1$ and $\eta$ is a small constant. We can see how this works in two cases:

\paragraph{Case 1: When trajectories have different accuracy results ($A_i^{\text{acc}} \neq 0$).} 
When some answers in a group are better than others, the accuracy signal should lead the training. Because we set $\delta < 1$, the efficiency penalty is always smaller than the accuracy reward. For a correct trajectory ($A_i^{\text{acc}} > 0$):
\begin{equation}
    \begin{aligned}
    A_i^{\text{CAS}} &\geq A_i^{\text{acc}} - (\delta |A_i^{\text{acc}}| + \eta) \\
    &= (1 - \delta) A_i^{\text{acc}} - \eta
    \end{aligned}
\end{equation}
Since $\delta < 1$ and $\eta$ is very small, $A_i^{\text{CAS}}$ stays positive. This ensures that correct answers are always rewarded, and the model won't get confused by the cost of using tools while it is still learning to be correct.

\paragraph{Case 2: When all trajectories have the same accuracy ($A_i^{\text{acc}} = 0$).} 
If every trajectory in a group gets the same result (for example, all are correct), the accuracy signal becomes zero and provides no direction. In this case, the optimization direction is decided entirely by the efficiency advantage:
\begin{equation}
    A_i^{\text{CAS}} = 0 + \text{clip}(A_i^{\text{eff}}, -\eta, \eta)
\end{equation}
Now, the efficiency score alone tells the model which path is better. It encourages the model to avoid wasting steps and find shorter reasoning paths. The small margin $\eta$ keeps these updates stable so the model doesn't change its behavior too drastically.

In summary, CAS acts as a safety mechanism: it stops efficiency costs from messing up the learning process when accuracy is the main goal (Case 1), but allows the efficiency signal to guide the model toward better reasoning paths once accuracy is mastered (Case 2).

The core objective of Clipped Advantage Shaping (CAS) is to prioritize correctness over efficiency. The formulation in Eq.~(11) is:
\begin{equation}
\begin{split}
    A_i^{\text{CAS}} &= A_i^{\text{acc}} + r_{\text{acc}, i} \cdot \beta \cdot \\
    &\quad \text{clip}(A_i^{\text{eff}}, -\delta |A_i^{\text{acc}}| - \eta, \delta |A_i^{\text{acc}}| + \eta)
\end{split}
\end{equation}
where $r_{\text{acc}, i} \in \{0, 1\}$ is the correctness mask, $\beta > 0$ is the efficiency weight, and $\delta \in [0, 1), \eta > 0$ are clipping parameters.

\paragraph{Scenario 1: Incorrect Trajectories ($r_{\text{acc}, i} = 0$).}
For incorrect responses, the mask zeros out the efficiency term:
$A_i^{\text{CAS}} = A_i^{\text{acc}}$.
Since $A_i^{\text{acc}}$ is typically negative for incorrect answers, CAS preserves the negative feedback strictly.

\paragraph{Scenario 2: Correct Trajectories ($r_{\text{acc}, i} = 1$).}
Here, the sign preservation depends on the group distribution:

\begin{itemize}
    \item \textbf{Case A: Mixed Group ($A_i^{\text{acc}} > 0$).} When the group contains both correct and incorrect responses, correct trajectories typically yield positive accuracy advantages. The worst-case advantage occurs when the efficiency penalty is maximal:
    \begin{align}
        A_i^{\text{CAS}} &\ge A_i^{\text{acc}} + \beta \cdot (-\delta |A_i^{\text{acc}}| - \eta) \\
        &= (1 - \beta \delta) A_i^{\text{acc}} - \beta \eta
    \end{align}
    \textbf{Numerical Guarantee:} To ensure $A_i^{\text{CAS}} > 0$, we require $A_i^{\text{acc}} > \frac{\beta \eta}{1 - \beta \delta}$. 
    With our default settings ($\beta=0.9, \delta=0.99, \eta=1\text{e-}4$), the denominator $1 - \beta\delta \approx 0.109$, and the required threshold is merely $A_i^{\text{acc}} \approx 8.2\text{e-}4$. 
    Since normalized advantages for correct answers in a mixed group are typically significant (e.g., $>0.1$), sign reversal is effectively prevented in practice.
    
    \item \textbf{Case B: All-Correct Group ($A_i^{\text{acc}} = 0$).} If all trajectories in a group are correct, $A_i^{\text{acc}} = 0$. In this convergence state, the clipping range shrinks to $[-\eta, \eta]$:
    \begin{equation}
        A_i^{\text{CAS}} = 0 + \beta \cdot \text{clip}(A_i^{\text{eff}}, -\eta, \eta) \ge -\beta\eta
    \end{equation}
    Here, the advantage is strictly bounded by $-\beta\eta$ (approx. $-9\text{e-}5$). This negligible magnitude ensures that while efficiency dictates the preference ranking among correct answers, it does not introduce destabilizing large negative gradients.
\end{itemize}

\begin{figure*}[t]
  \centering
  \small
  \begin{tcolorbox}[
      colback=white, 
      colframe=gray!80!black, 
      arc=2mm, 
      boxrule=0.8pt,
      title=\textbf{Case Study: Cognitive Offloading in simple Tasks},
      fonttitle=\bfseries\large,
      coltitle=white,
      colbacktitle=gray!80!black,
      left=4mm, right=4mm, top=4mm, bottom=4mm,
      middle=0mm 
  ]
  
      \textbf{Question:} A candle melts by 2 centimeters every hour that it burns. How many centimeters shorter will a candle be after burning from 1:00 PM to 5:00 PM?
      
      \tcbline 
  
      \begin{tcolorbox}[
          colback=red!2!white, colframe=red!60!black, arc=1mm,
          title=Baseline: Inefficiency via Cognitive Offloading,
          fonttitle=\bfseries
      ]
      \textbf{Thought Trace:} 
      Alright, let's tackle this problem step by step. So, the problem states that a candle melts by 2 centimeters every hour it burns. We need to find out how many centimeters shorter the candle will be after burning from 1:00 PM to 5:00 PM. [...] So, if per hour it melts 2 cm, then in 4 hours, it would melt 2 cm/hour * 4 hours = 8 cm. Wait, let me verify that. Let’s use code to calculate the total melting.
  
      \begin{lstlisting}[language=Python, basicstyle=\ttfamily\tiny, backgroundcolor=\color{gray!10}, breaklines=true]
# Redundant tool call for simple multiplication: 4 * 2
hours = 5 - 1  # From 1:00 PM to 5:00 PM inclusive
melting_per_hour = 2  # cm
total_melting = hours * melting_per_hour
print(total_melting)
      \end{lstlisting}
      \textbf{Response:} \texttt{8}
  
      The code execution result confirms that the total melting is 8 cm. So, the candle will be 8 centimeters shorter.
  
      \tcbline
      \textbf{Tool Calls:} 1 \hfill \textbf{Final Answer:} \boxed{8} \textcolor{green!60!black}{\ding{52}} 
      \end{tcolorbox}

      \begin{tcolorbox}[
          colback=green!2!white, colframe=green!60!black, arc=1mm,
          title=Ours: Efficiency via Internal Reasoning,
          fonttitle=\bfseries
      ]
      \textbf{Thought Trace:} 
      Okay, let me try to solve this problem. Hmm, so a candle melts by 2 centimeters every hour that it burns. [...] The question is asking: How many centimeters shorter will a candle be after burning from 1:00 PM to 5:00 PM? Hmm, so that's a 4-hour period, right? From 1:00 PM to 5:00 PM is 4 hours. So if the candle melts at a rate of 2 centimeters per hour, then over 4 hours, it would melt 2 cm/hour * 4 hours = 8 centimeters. So the candle would be 8 centimeters shorter after 4 hours. Hmm, that seems straightforward. I don't think I missed anything here.
  
      \tcbline
      \textbf{Tool Calls:} 0 \hfill \textbf{Final Answer:} \boxed{8} \textcolor{green!60!black}{\ding{52}} 
      \end{tcolorbox}
  
  \end{tcolorbox}
  
  \caption{Comparison of reasoning efficiency under redundant tool use. Results show our model maintains robust inference via internal logic, effectively avoiding the unnecessary cognitive offloading and efficiency loss seen in the baseline.}
  \label{fig:candle_logic_comparison}
\end{figure*}

\begin{figure*}[t]
  \centering
  \small
  \begin{tcolorbox}[
      colback=white, 
      colframe=gray!80!black, 
      arc=2mm, 
      boxrule=0.8pt,
      title=\textbf{Case Study: Reasoning Impairment under Tool-Induced Noise},
      fonttitle=\bfseries\large,
      coltitle=white,
      colbacktitle=gray!80!black,
      left=4mm, right=4mm, top=4mm, bottom=4mm,
      middle=0mm 
  ]
  
      \textbf{Question:} Find the number of ways to place a digit in each cell of a 2x3 grid so that the sum of the two numbers formed by reading left to right is $999$, and the sum of the three numbers formed by reading top to bottom is $99$. The grid below is an example of such an arrangement because $8+991=999$ and $9+9+81=99$.

      \tcbline 
    \begin{tcolorbox}[
      colback=red!2!white, colframe=red!60!black, arc=1mm,
      title=Baseline: Logic Erosion via Repeated Tool-Execution Noise,
      fonttitle=\bfseries
    ]
    \textbf{Thought Trace:} 
    Alright, let's tackle this problem step by step. [...] Maybe the three numbers formed by reading top to bottom are the first column sum, second column sum, and third column sum. Then, their sum is 99. [...] I can use code to iterate through all possible values and count the valid grids.

    \begin{lstlisting}[language=Python, basicstyle=\ttfamily\tiny, backgroundcolor=\color{gray!10}, breaklines=true]
    # First Attempt: Brute-forcing with column sum logic (a+d+b+e+c+f == 99)
    count = 0
    for a in range(1, 10):
    for b in range(0, 10):
      for c in range(0, 10):
          for d in range(1, 10):
              for e in range(0, 10):
                  for f in range(0, 10):
                      if (100*a+10*b+c) + (100*d+10*e+f) == 999:
                          if (a+d) + (b+e) + (c+f) == 99: count += 1
    print(count)
    \end{lstlisting}
    \textbf{Response:} \texttt{0}

    Hmm, the code execution result is 0. That suggests there might be an issue with the logic. [...] Let me adjust the code to print out the valid grids so I can verify.
    \begin{lstlisting}[language=Python, basicstyle=\ttfamily\tiny, backgroundcolor=\color{gray!10}, breaklines=true]
    valid_grids = []
    for a, b, c, d, e, f in itertools.product(range(1,10), range(10), ...):
    if sum_lr == 999 and col_sum == 99: valid_grids.append((a,b,c,d,e,f))
    print(len(valid_grids))
    \end{lstlisting}
    \textbf{Response:} \texttt{0}

    The code execution result is still 0. [...] Maybe the three numbers formed by reading top to bottom are the first column sum, second column sum, and third column sum. [...] Wait, maybe I should write code to generate all possible grids (digits 0-9 for each cell):
    \begin{lstlisting}[language=Python, basicstyle=\ttfamily\tiny, backgroundcolor=\color{gray!10}, breaklines=true]
    # Third Attempt: Repeating the same flawed logic with minor cell range adjustments
    count = 0
    for a in range(1, 10):
    # ... (similar nested loops)
                      if num1 + num2 == 999 and col_sum == 99: count += 1
    print(count)
    \end{lstlisting}
    \textbf{Response:} \texttt{0}

    The code execution result is 0. This means there are no valid grids that satisfy both conditions. Therefore, the answer is 0.

    \tcbline
    \textbf{Tool Calls:} 3 \hfill \textbf{Final Answer:} \boxed{0} \textcolor{red!80!black}{\ding{56}}
    \end{tcolorbox}

    \begin{tcolorbox}[
      colback=green!2!white, colframe=green!60!black, arc=1mm,
      title=Ours: Precise Formalization for Reliable Reasoning,
      fonttitle=\bfseries
    ]
    \textbf{Thought Trace:} 
    [...] Reading top to bottom, the three numbers would be (10a + d), (10b + e), and (10c + f). Their sum should be 99. [...] This seems manageable. Let me use code to iterate through all possible digit assignments (0-9) for the six variables and count the number of valid solutions.

    \begin{lstlisting}[language=Python, basicstyle=\ttfamily\tiny, backgroundcolor=\color{gray!10}, breaklines=true]
    import itertools
    valid_count = 0
    for a, b, c, d, e, f in itertools.product(range(10), repeat=6):
    sum_lr = (100 * a + 10 * b + c) + (100 * d + 10 * e + f)
    sum_tb = (10 * a + d) + (10 * b + e) + (10 * c + f)
    if sum_lr == 999 and sum_tb == 99: valid_count += 1
    print(valid_count)
    \end{lstlisting}
    \textbf{Response:} \texttt{45}

    The code returns 45 as the number of valid grids. Therefore, the number of ways is 45.

    \tcbline
    \textbf{Tool Calls:} 1 \hfill \textbf{Final Answer:} \boxed{45} \textcolor{green!60!black}{\ding{52}} 
    \end{tcolorbox}

    \end{tcolorbox}

    \caption{Comparison of reasoning under tool-induced noise. Our model maintains reliability via accurate formalization, mitigating the noise-driven reasoning impairment seen in the baseline's failure spiral.}
    \label{fig:formalization_comparison}
    \end{figure*}

    \begin{table*}[ht]
      \centering
      \small

      \begin{paperbox}{An example from the AIME dataset}
          \textbf{Question:} Consider the paths of length $16$ that follow the lines from the lower left corner to the upper right corner on an $8\times 8$ grid. Find the number of such paths that change direction exactly four times.
          
          \hrulefill 
          
          \noindent \textit{Alright, so I need to find the number of paths of length 16 on an 8x8 grid... (0,0) to (7,7). Each path is of length 16. If we let the number of R steps be $r$ and U steps be $u$, then $r+u=16$ and $r-u=7$. Solving these: $r=11.5, u=4.5$. Wait, that can't be. Steps have to be integers. So, maybe the problem is not about displacement? Wait, to get from (0,0) to (7,7), we need exactly 7 R steps and 7 U steps. That's 14 steps. But the problem says length 16. Maybe I should use code to generate all possible sequences of 16 steps and check direction changes.}
      
          \begin{lstlisting}
      # Tool Call 1: Enumerating all 16-step sequences
      from itertools import product
      def count_direction_changes(path):
          directions = []
          # (Simplified logic that incorrectly handles direction vector list)
          # ... logic omitted for brevity ...
          return len(directions)
      
      valid_paths = 0
      for path in product(['R', 'U'], repeat=16):
          if count_direction_changes(path) == 4:
              valid_paths += 1
      print(valid_paths)
          \end{lstlisting}
          \noindent \texttt{>> Output: 0}
      
          \noindent \textit{Hmm, the result is 0, which suggests the logic might be incorrect. Let me revise. I should track switches by checking consecutive steps. If the current step is different from the previous one, it's a direction change.}
      
          \begin{lstlisting}
      # Tool Call 2: Revised counting logic
      from itertools import product
      def count_direction_changes(path):
          changes = 0
          prev = None
          for step in path:
              if prev is None:
                  prev = step
                  continue
              if step != prev:
                  changes += 1
              prev = step
          return changes
      
      valid_paths = 0
      for path in product(['R', 'U'], repeat=16):
          if count_direction_changes(path) == 4:
              valid_paths += 1
      print(valid_paths)
          \end{lstlisting}
          \noindent \texttt{>> Output: 2730}
      
          \noindent \textit{The code execution result is 2730, which matches the known answer. Therefore, the number of paths is 2730.}
      
      \end{paperbox}
      \caption{An example of a \textbf{Reasoning Failure} on a problem from AIME. }
      \label{tab:case_reason_failure} 
      \end{table*}

      \begin{table*}[ht]
        \centering
        \small

        \begin{paperbox}{An example from the AIME dataset.}
            \textbf{Question:} Alice and Bob play the following game. A stack of $n$~tokens lies before them. The players take turns with Alice going first. On each turn, the player removes either $1$ token or $4$ tokens from the stack. Whoever removes the last token wins. Find the number of positive integers $n$ less than or equal to $2024$ for which there exists a strategy for Bob that guarantees that Bob will win the game regardless of Alice's play.
        
            \hrulefill \\ \medskip
            \textbf{Model Output:} \\
        
            \textit{Alice and Bob are playing a game... removes the last token wins. This is a classic two-player game theory problem. Let $f(n)$ be True if the position is a winning position for the player about to move. $f(0) = \text{False}$ because the game is already over.} \\
        
            \textit{For $n=1$, take 1 token to leave 0, so $f(1) = \text{True}$. For $n=2$, can only take 1 to leave 1, which is a winning position for the opponent, so $f(2) = \text{False}$. For $n=3$, the player can take 1 token, leaving 2. The opponent can then take 1, leaving 1, which is a winning position for the opponent. Therefore, $f(3) = \text{False}$. (Note: This is an error, $f(3)$ can reach $f(2)=L$ and should be $W$). For $n=4$, take 4 tokens to leave 0, so $f(4) = \text{True}$.} \\
        
            \textit{Continuing this pattern: $n=5$ (lose), $n=6$ (win), $n=7$ (win), $n=8$ (lose), $n=9$ (lose), $n=10$ (win). The pattern seems to be that positions where $n \mod 5 = 0$ are losing positions. Let me verify... yes, all multiples of 5 are losing positions. Therefore, Bob wins if the starting $n$ is a multiple of 5.} \\
        
            \textit{Total numbers from 1 to 2024 is 2024. The number of multiples of 5 is $\lfloor 2024/5 \rfloor = 404$. Therefore, the number of non-multiples of 5 is $2024 - 404 = 1620$.}
        
        \end{paperbox}
        \caption{An example of a \textbf{Missing Tool-use} error on a problem from AIME}
        \label{tab:case_no_tool_failure}
        \end{table*}

          \begin{table*}[h!]
            \centering
            \small
          
            \begin{paperbox}{Template Prompt for RL Training}
              \texttt{Solve the following problem step by step. You now have the ability to selectively write executable Python code to enhance your reasoning process. The Python code will be executed by an external sandbox, and the output (wrapped in <interpreter>output\_str</interpreter>) can be returned to aid your reasoning and help you arrive at the final answer. The Python code should be complete scripts, including necessary imports.} \\
              \texttt{ } \\
              \texttt{Each code snippet is wrapped with <code>\textbackslash n{\textasciigrave}{\textasciigrave}{\textasciigrave}python\textbackslash ncode snippet\textbackslash n{\textasciigrave}{\textasciigrave}{\textasciigrave}\textbackslash n</code>.} \\
              \texttt{The last part of your response should be in the following format:} \\
              \texttt{<answer>} \\
              \texttt{\textbackslash boxed\{'The final answer goes here.'\}} \\
              \texttt{</answer>} \\
              \texttt{ } \\
              \texttt{*user question:*} \\
              \texttt{\textbraceleft question\textbraceright} \\
              \texttt{ } \\
              \texttt{Remember to place the final answer in the last part using the format:} \\
              \texttt{<answer>} \\
              \texttt{\textbackslash boxed\{'The final answer goes here.'\}} \\
              \texttt{</answer>}
            \end{paperbox}
          
            \caption{The full prompt template following the ReTool protocol \cite{retool}. This template is used consistently during both the SFT and RL phases to maintain reasoning consistency.}
            \label{tab:prompt_template_full}
          \end{table*}

\end{document}